\documentclass[pdflatex,sn-mathphys-num]{sn-jnl}

\usepackage{array}
\usepackage{inconsolata}
\usepackage{pgfplots}
\usepackage{pgfplotstable}
\pgfplotsset{compat=1.17}
\usepackage{multirow}
\usepackage{colortbl}
\usepackage{xcolor}
\usepackage{lipsum}
\usepackage{float}
\usepackage{subfig}
\usepackage{graphicx}%
\usepackage{multirow}%
\usepackage{amsmath,amssymb,amsfonts}%
\usepackage{amsthm}%
\usepackage{mathrsfs}%
\usepackage[title]{appendix}%
\usepackage{soul}
\usepackage{textcomp}%
\usepackage{manyfoot}%
\usepackage{booktabs}%
\usepackage{algorithm}%
\usepackage{algorithmicx}%
\usepackage{algpseudocode}%
\usepackage{listings}%
\usepackage[utf8]{inputenc}
\usepackage{verbatim}
\UseRawInputEncoding
\usepackage{tikz}
\usetikzlibrary{arrows.meta}
\usepackage{adjustbox}
\usepackage{tabularx}
\usepackage{booktabs}
\usepackage{tikz}
\usepackage{footmisc}
\usepackage{url}
\usepackage{listings}
\usepackage[dvipsnames]{xcolor}
\usepackage{pifont}  
\usepackage{xcolor}  
\definecolor{skyblue}{RGB}{135, 206, 235}  
\usepackage{orcidlink}
\usetikzlibrary{shapes.geometric, arrows, positioning}

\hypersetup{
	colorlinks=true,
	linkcolor=blue,
	filecolor=blue,      
	urlcolor=blue,
	citecolor = blue
}

\begin{document}

\title[{Mixture of Detectors: A Compact View of Machine-Generated Text Detection}]{Mixture of Detectors: A Compact View of Machine-Generated Text Detection}

\author[1]{\fnm{Sai Teja} \sur{Lekkala}}\email{lekkalad\_ug\_22@cse.nits.ac.in}
\author[1]{\fnm{Yadagiri} \sur{Annepaka}}\email{annepaka22\_rs@cse.nits.ac.in}
\author[2]{\fnm{Arun Kumar} \sur{Challa}}\email{challaa\_ug\_22@ee.nits.ac.in}
\author[1]{\fnm{Samatha Reddy} \sur{Machireddy}}\email{machireddys\_ug\_22@cse.nits.ac.in}
\author*[1]{\fnm{Partha} \sur{Pakray}}\email{partha@cse.nits.ac.in}
\author[1]{\fnm{Chukhu} \sur{Chunka}}\email{chukhu@cse.nits.ac.in}

\affil[1] {\orgdiv{Computer Science \& Engineering, NIT Silchar}, \country{India}}
\affil[2] {\orgdiv{Electrical Engineering, NIT Silchar}, \country{India}}

\abstract{Large Language Models (LLMs) are gearing up to surpass human creativity. The veracity of the statement needs careful consideration. In recent developments, critical questions arise regarding the authenticity of human work and the preservation of their creativity and innovative abilities. This paper investigates such issues. This paper addresses machine-generated text detection across several scenarios, including document-level binary and multiclass classification or generator attribution, sentence-level segmentation to differentiate between human-AI collaborative text, and adversarial attacks aimed at reducing the detectability of machine-generated text. We introduce a new work called BMAS English: an English language dataset for binary classification of human and machine text, for multiclass classification, which not only identifies machine-generated text but can also try to determine its generator, and Adversarial attack addressing where it is a common act for the mitigation of detection, and Sentence-level segmentation, for predicting the boundaries between human and machine-generated text. We believe that this paper will address previous work in Machine-Generated Text Detection (MGTD)  in a more meaningful way.}


\maketitle

\section{Introduction}\label{Introduction}
The rapid emergence of Large Language Models (LLMs) \cite{chang2024survey,annepaka2025large} such as ChatGPT, Grok, and DeepSeek marks a transformative shift in automated text generation. These models, built upon the Transformer architecture introduced by \textit{``Attention Is All You Need''} \cite{vaswani2017attention}, demonstrate exceptional capabilities in producing coherent, contextually relevant, and human-like text. Since their inception, LLMs have evolved through major milestones-ranging from the early Generative Pre-trained Transformers (GPT-1) to GPT-5 and the o-series, and extending to state-of-the-art (SOTA) architectures such as DeepSeek-V3, R1 \cite{guo2025deepseek}, Sparrow, Anthropic’s Claude Sonnet 3.5 and 3.7, and Grok 3 and 4 from Explainable AI (X-AI). The widespread accessibility of these models has revolutionized productivity across journalism, social media, education, academic writing, and beyond. They have also integrated advanced techniques, including Reinforcement Learning from Human Feedback (RLHF), Mixture-of-Experts architectures, and multi-step reasoning, with some high-performing variants even being released as open-source. While these advances offer immense potential for accelerating innovation and democratizing AI technologies, they also raise pressing concerns about the erosion of work authenticity and authorship attribution. As LLM outputs become increasingly indistinguishable from human-authored content, the boundary between human and AI authorship is blurring at an unprecedented pace. This growing ambiguity has significant implications for academic integrity, intellectual property, and societal trust in information systems. Without effective detection mechanisms, it may become impossible to verify whether a given work, such as a research article, business proposal, or educational material originates from a human author, an AI system, or a collaborative process between the two. Consequently, AI-generated text detection has emerged as a critical area of research, with the aim of protecting authenticity, ensuring transparency, and minimizing the potential misuse of generative AI technologies.

\begin{figure*}[htbp]
\centering
\includegraphics[width=1.0\linewidth]{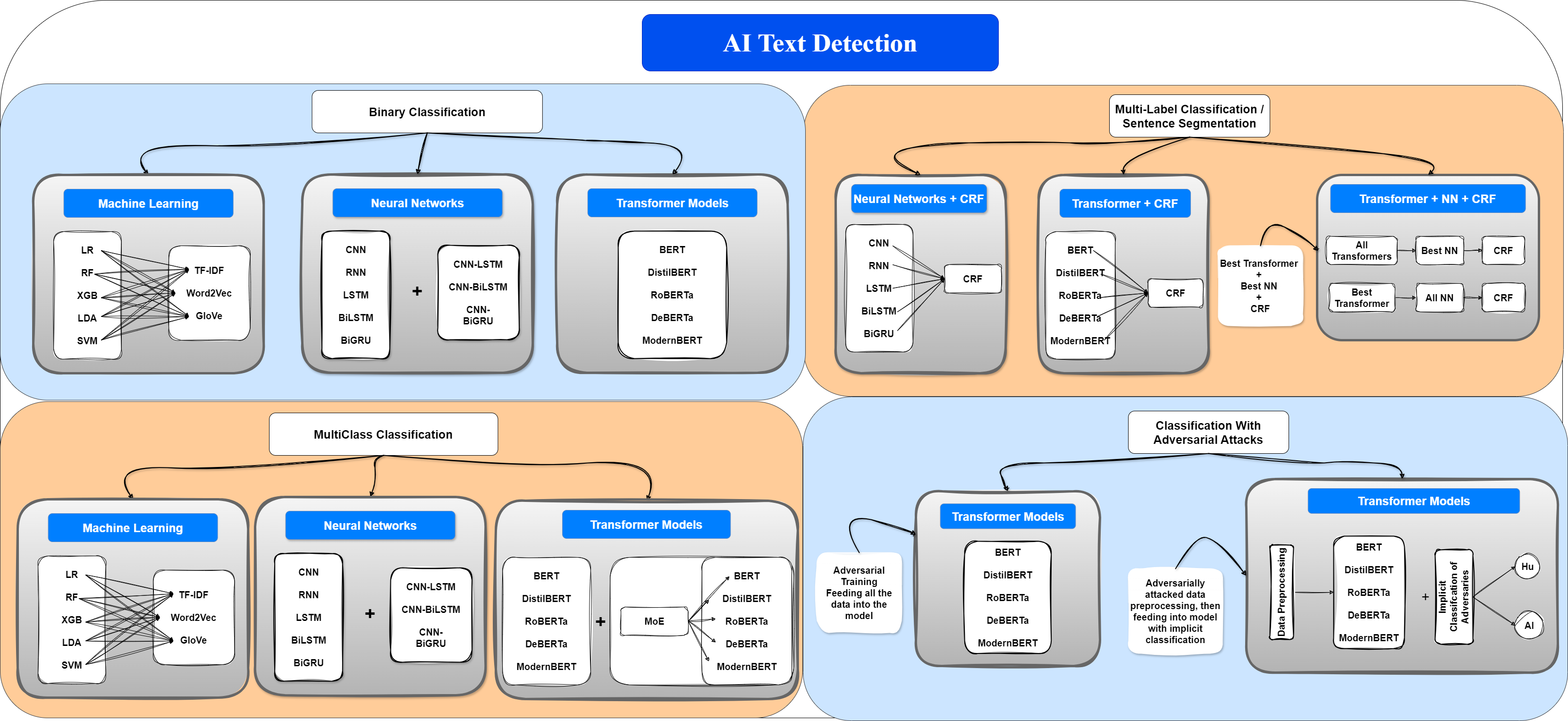} 
\caption{All Experiments for the Classification, Segmentation, and Adversarial Attacks addressing.}
\label{AllExperiments}
\end{figure*}

\section{Problem Formulation and Research Scope}
As outlined in the abstract and illustrated in Figure~\ref{AllExperiments}, we address the machine-generated text detection problem through three complementary approaches. For each approach, we construct dedicated datasets and train multiple models to enable comparative analysis, experimental evaluation, and benchmarking. The three problem settings are as follows:

\textbf{1) Binary and Multiclass Classification:}  
We focus on English text and curate a corpus of human-written and LLM-generated samples across key domains, enabling (a) \textit{binary classification} to distinguish human-written vs. AI-generated text, and (b) \textit{multiclass classification} to further identify the generating LLM.

\textbf{2) Adversarial Robustness:}  
Existing detectors are often vulnerable to adversarial modifications, where AI-generated text can be manipulated to evade detection through techniques such as synonym substitution, homoglyph replacement, misspellings, character insertion or deletion, and paraphrasing. To address this, we introduce a dataset that incorporates five common adversarial attack types that are frequently used to bypass detection systems, enabling the development of more robust models.

\textbf{3) Mixed Text Boundary Detection:}  
While the first two tasks operate at the document level, real-world scenarios often involve \textit{mixed-authorship} text, where human-written and AI-generated segments are interleaved. To capture this complexity, we construct a mixed-text dataset comprising three variants: (1) Human-written text followed by AI continuation, (2) AI-generated text followed by human continuation, and (3) Fully interleaved human–AI segments. Each sample is annotated with precise boundary labels: \textit{Human End Boundary}, \textit{Machine End Boundary}, and the exact word-index where authorship transitions occur.

\section{Related Work} \label{rel_work}
The detection of Machine-Generated Text (MGT) detection has been predominantly framed as a binary classification task \cite{zellers2019defending,gehrmann2019gltr}. Existing methodologies for MGT detection can be broadly classified into supervised and unsupervised approaches. Supervised methods \cite{wang2023m4,uchendu2021turingbench} utilize labeled datasets to train discriminative models for classification. In contrast, unsupervised techniques rely on intrinsic textual features such as perplexity, log-rank statistics \cite{mitchell2023detectgpt,hans2024spotting}, or leverage watermarking schemes \cite{kirchenbauer2023watermark,zhao2023protecting} to distinguish machine-generated content. In this work, we focus primarily on supervised detection approaches, given their demonstrated effectiveness in using annotated data to improve classification performance. \cite{wang2023m4} evaluates several supervised detectors, such as RoBERTa \cite{liu2019roberta}, XLM-R \cite{conneau2019unsupervised}, a logistic regression classifier with Word2Vec features \cite{gehrmann2019gltr}, models leveraging stylistic features \cite{li2014authorship}, and classifiers using NELA features \cite{horne2019robust}. Similar analyses of supervised methods have been conducted in recent work \cite{guo2023close,xiong2024fine}. Machine-Generated Text Detection (MGTD) has been the work of many individuals who have been exploring this since the evolution of the LLMs. We are inspired by the work of M4 \cite{wang2024m4gt}, MAGE \cite{li2023mage}, and RAID \cite{dugan2024raid}, as they created a large and wide corpus for this MGTD. M4 is multilingual, MAGE has multidomain, and both M4 and MAGE have several LLMs and data from RAID with several types of adversarial attacks.  We explored these tasks by training from traditional Machine Learning (ML) classifiers to neural network models, and then pre-trained transformer models, and their other methodologies, with them like the inclusion of Linguistic feature layers to the Transformer output layers, and addressing adversarial attacks, also a methodology from a shared task paper by \cite{lekkala-etal-2025-cnlp}. For the Sentence Segmentations, the works of SeqXGPT \cite{wang-etal-2023-seqxgpt}, Real Or Fake Text (RoFT) \cite{dugan-etal-2020-roft}, RoFT-ChatGPT \cite{kushnareva2023ai} and \cite{zeng2024towards}. All the major works mentioned above have undergone multiple progressive phases of development, including extensions to multiclass classification for both generator and domain identification, as well as adaptations for multilingual detection scenarios.  Furthermore, considerable effort has been devoted to incorporating a diverse range of adversarial attacks, aiming to rigorously challenge model robustness under varied threat conditions. In line with this progression, our work is committed to constructing a comprehensive dataset for all scenarios in MGTD that systematically addresses all conceivable cases within the MGTD framework.

\subsection{Our Key Contributions}
\noindent This work introduces an innovative hybrid architecture that combines a segmentation-based token-level classification model with adversarial robustness mechanisms, enabling accurate boundary detection between human and AI-generated text. The proposed method is evaluated across multiple syntactic attack scenarios, demonstrating superior segmentation precision and stability under challenging conditions.
\begin{enumerate}
    \item \textbf{Introduction of the English-BMAS Datasets:} A comprehensive benchmark\footnote{\url{https://huggingface.co/datasets/saiteja33/BMAS}} for mixed AI text detection in several scenarios in English, incorporating diverse domains, collaboration scenarios, and writing styles.
    
    \item \textbf{Extensive Experimental Evaluation:} Large-scale experiments conducted across multiple datasets, model architectures, and evaluation metrics to ensure robust and generalizable findings\footnote{\url{https://github.com/saitejalekkala33/E-BMAS-A-mixture-of-AI-Detectors.git}}.
    
    \item \textbf{Novel HardMoE and SoftMoE Detection Architectures:} Two complementary mixture-of-experts–based models designed to capture both explicit and subtle cues of AI-generated text.
    
    \item \textbf{Implicit Adversarial Detection Framework:} A new detection paradigm targeting adversarially modified text, including syntactic and stylistic perturbations that evade traditional classifiers.
    
    \item \textbf{Sentence-Level Segmentation and Boundary Detection:} An advanced segmentation module to identify precise authorship transition points at sentence granularity.
\end{enumerate}

The rest of this paper is organized as follows. In Section \ref{sec:datasetdescription}, we detailed the creation of an enhanced benchmark dataset. Section \ref{sec:methodology} outlines the way of experimentation in detecting AI-generated text in various scenarios. The experimental setup is described in Section \ref{sec:experimentalsetup}, followed by the corresponding experimental results in Section \ref{sec:res_discussion}. In the section \ref{sec:ablation}, we presented the Ablation Study of experiments with zero-shot methods and other variations in the model. In conclusion, Section \ref{sec:conclusion} presents the key findings of the study and suggests potential avenues for future research, then in Section \ref{sec:limitations}, we presented a few limitations of the work, which we bound to do them in the future.

\section{Dataset descriptions}\label{sec:datasetdescription}
\subsection{BMAS Dataset}
\begin{figure*}[htbp]
\centering
\includegraphics[width=0.8\linewidth]{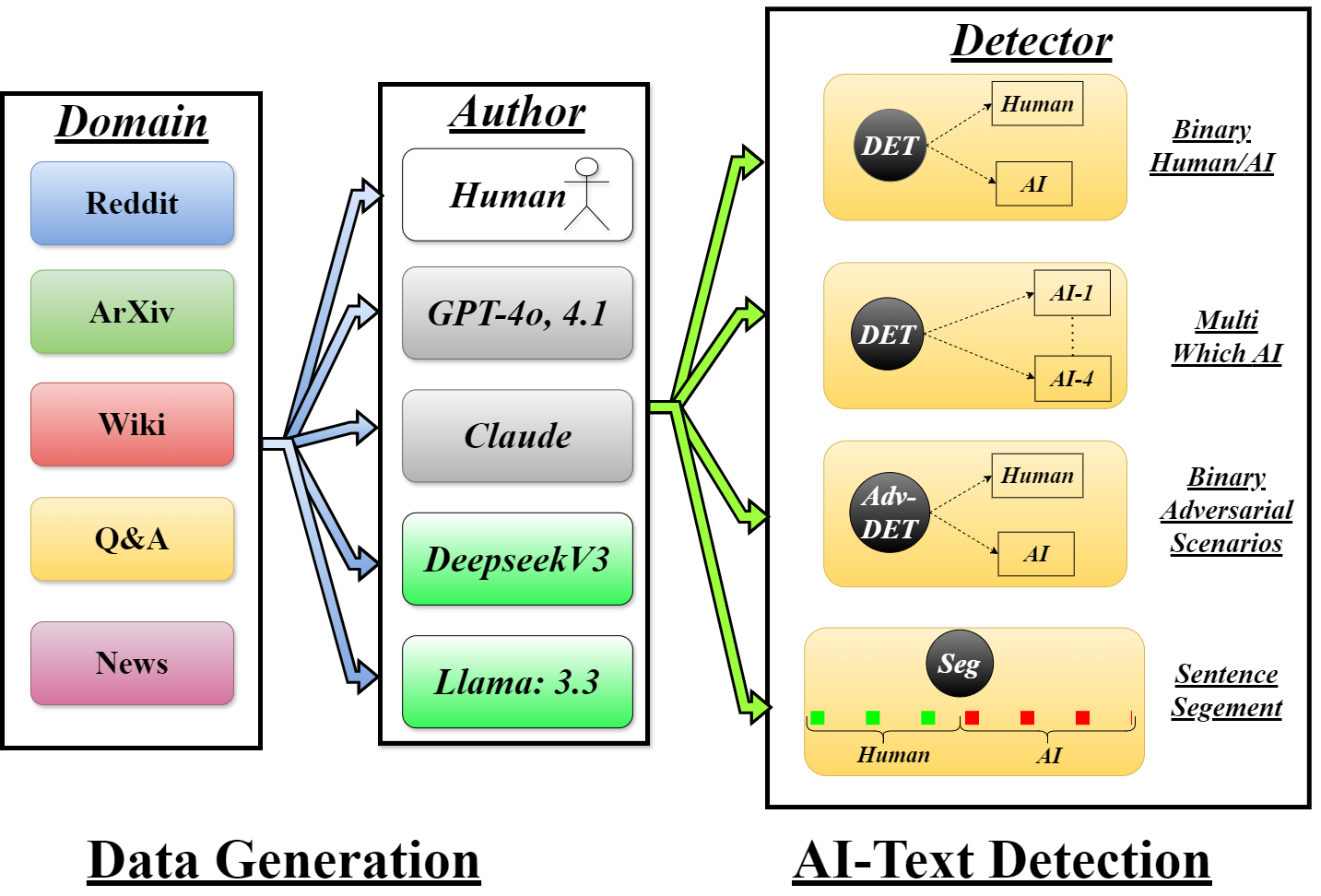} 
\caption{Text data generation from diverse domains using AI models, where blocks shaded in \textcolor{gray}{gray} represent \textcolor{gray}{closed-source} models and those in \textcolor{green}{green} represent \textcolor{green}{open-source} models. The figure also illustrates the detection of AI-generated text in various scenarios.}
\label{fig:datagen}
\end{figure*}

We curated a diverse dataset comprising both human-written and AI-generated texts from five widely used domains in real-world applications: reddit, news articles, wikipedia content, scientific abstracts from arXiv, and general-purpose question answering (Q\&A). Human-authored texts were primarily sourced from the MAGE and M4 datasets, which provide comprehensive coverage for most of these domains. For the news domain, we utilized the XSUM \cite{xsum-emnlp} dataset. Our dataset is designed to be simple, concise, robust, and easy to use, with carefully balanced domain coverage and controlled distributions across different data types. Detailed statistics regarding domain-wise and source-wise distribution are provided in Tables~\ref{tab:data_distribution_domain_wise}, and~\ref{tab:mixed_text_data}, the data generation from the various sources and AI models is illustrated in Fig~\ref{fig:datagen}.

\begin{enumerate}
    \item \textbf{Binary Classification Dataset:} Contains 80,000 samples equally split between human-written (40,000) and AI-generated (40,000) texts. The data is divided into 70\%, 20\%, 10\% splits for train, validation, and test.
    \item \textbf{Multiclass Classification Dataset:} 80,000 samples with the label distribution as 40,000 human, and 10,000 each from four language models, OpenAI, Claude, Llama, and DeepSeek. The dataset is split in a 70/20/10 ratio for training, validation, and testing, respectively.
    \item \textbf{Adversarial Detection Dataset:} We apply five perturbation strategies to the original 80,000 sample dataset, generating six versions per sample (5 attacked + 1 original), resulting in 480,000 total samples. The attacks are, 1) Synonym Substitution, 2) Misspelling, 3) Homoglyph Replacement, 4) Upper-Lower Swap, and 5) Zero-Width Space Insertion. The same 70/20/10 split is used.
    \item \textbf{Sentence-Level Segmentation Dataset:} 46,830 mixed-authorship samples labeled at the sentence level. This includes 20,000 samples with H$\rightarrow$M, 16,995 samples M$\rightarrow$H, and 9,875 samples with complex human-AI mixing.
\end{enumerate}

\begin{table}[ht]
\centering
\caption{Domain-wise distribution of human and LLM-generated corpus. Values represent non-adversarial samples, with numbers in parentheses denoting $(5$ adversarial variants $+1$ original$)$.}
\label{tab:data_distribution_domain_wise}
\begin{tabular}{lccccc}
\toprule
\textbf{Model} & \textbf{Reddit} & \textbf{News} & \textbf{Wiki} & \textbf{Arxiv} & \textbf{Q\&A} \\
\midrule
\textbf{Human}     & 10k (5+1) & 10k (5+1) & 10k (5+1) & 10k (5+1) & 10k (5+1) \\
\textbf{GPT-4o}    & 2k (5+1)  & 2k (5+1)  & 2k (5+1)  & 2k (5+1)  & 2k (5+1)  \\
\textbf{Claude-3.5} & 2k (5+1)  & 2k (5+1)  & 2k (5+1)  & 2k (5+1)  & 2k (5+1)  \\
\textbf{LLamA3.3:70b}     & 2k (5+1)  & 2k (5+1)  & 2k (5+1)  & 2k (5+1)  & 2k (5+1)  \\
\textbf{DeepSeek-V3:671b}  & 2k (5+1)  & 2k (5+1)  & 2k (5+1)  & 2k (5+1)  & 2k (5+1)  \\
\bottomrule
\end{tabular}

\end{table}

\begin{table}[ht]
\centering
\caption{Distribution of Human-AI Collaborative Texts by Interaction Type: 'HM' (Human-initiated, Machine-ended), 'MH' (Machine-initiated, Human-ended), and 'Mix' (Multiple alternating contributions with high coherence)}
\label{tab:mixed_text_data}
\scriptsize
\begin{tabular}{l c c c c c}
\toprule
\textbf{Model} & \textbf{Reddit} & \textbf{News} & \textbf{Wiki} & \textbf{ArXiv} & \textbf{Q\&A} \\
\midrule
GPT-4o(\textbf{\textit{HM}})          & 2k   & 2k   & 2k   & 2k   & 2k   \\
DeepseekV3(\textbf{\textit{HM}})        & 2k   & 2k   & 2k   & 2k   & 2k   \\
\midrule
GPT-4o(\textbf{\textit{MH}})          & 2k   & 2k   & 2k   & 2k   & 2k   \\
DeepseekV3(\textbf{\textit{MH}})        & 957  & 1998 & -    & 2k   & 2k   \\
\midrule
GPT-4.1-Mini(\textbf{\textit{Mix}})   & 986  & 1k   & 981  & 998  & 971  \\
GPT-4.1(\textbf{\textit{Mix}})        & 987  & 1k   & 984  & 998  & 970  \\
\bottomrule
\end{tabular}
\end{table}

\section{Methodology}\label{sec:methodology}
\subsection{Detectors}\label{sec:detectors}
We perform a thorough suite of experiments that span the evolution of binary-class, multi-class, binary-label classification methods from traditional ML approaches to the latest transformer-based architectures. We initially evaluated classical models for baseline binary and multi-class classification and introduced a new method that uses an expert routing system among the transformer models for the classification. Building upon this, we explore the capabilities of transformer models under both non-adversarial and adversarial settings. For adversarial robustness by training and evaluate models under adversarial conditions. We employ two approaches: adversarial training, where models are explicitly trained on perturbed inputs, and implicit adversarial detection, where the architecture is designed to be inherently sensitive to adversarial shifts. For boundary detection tasks, we incorporate Conditional Random Field (CRF) \cite{zheng2015conditional} layers on top of transformer-based encoders and neural networks.

\subsection{Binary and Multi-class Classification}
\subsubsection{Machine Learning Classifiers}
We selected the 5 best-performing traditional classifiers for analysis (taken these into consideration upon our prior works): Logistic Regression (\textit{LR}) \cite{lavalley2008logistic}, Random Forest (\textit{RF}) \cite{pal2005random}, Extreme Gradient Boosting (\textit{XGB}) \cite{chen2015xgboost}, Linear Discriminant Analysis (\textit{LDA}) \cite{balakrishnama1998linear}, and Support Vector Machine (\textit{SVM}) \cite{cervantes2020comprehensive}. We employed these classifiers for both binary and multi-class classification tasks. Each classifier was evaluated using three distinct text embedding techniques: Term Frequency-Inverse Document Frequency (\textit{TF-IDF}) \cite{dessi2021tf}, Word2Vec \cite{allen2019analogies}, and NEws LAndscape (\textit{NELA}) embeddings \cite{horne2019robust}. In total, this resulted in 30 different experimental configurations across the selected classifiers and embedding methods.

\subsubsection{Neural Network Classifiers}
To evaluate the neural network models in detecting AI-text hybrid text in binary using a sigmoid activation and multiclass using a softmax activation, we experimented with a variety of baseline and hybrid architectures. The selected models include: \textit{Convolutional Neural Network (CNN) \cite{luan2019research}, Recurrent Neural Network (RNN) \cite{liu2016recurrent}, Long-short Term Memory (LSTM) \cite{zhou2015c}, Bi-directional Long Short-Term Memory (Bi-LSTM), Bidirectional Gated Recurrent Unit (BiGRU) \cite{tie2025research,hu2021bi}, CNN-LSTM, CNN-BiLSTM}, and \textit{CNN-BiGRU}. Across all architectures, we utilized \textit{TF-IDF} as the primary text embedding technique, which consistently outperformed other embeddings in ML models.

\begin{enumerate}
    \item \textbf{Non-Hybrid Neural Network Models}: The non-hybrid neural models taken are \textit{CNN, RNN, LSTM, BiLSTM}, and \textit{BiGRU}. The input texts into these models are vectorized using the \textit{TF-IDF} representation, as the TF-IDF vectorization method has high scores. Each main layer was followed by batch normalization and regularized using a dropout rate of 0.3. The resulting representations were flattened and passed through a dense classification head consisting of two fully connected layers with 128 and 64 units, activated using ReLU and regularized with dropout rates of 0.5 and 0.3.
    \item \textbf{Hybrid Neural Network Models}: The hybrid models considered for experimentation are \textit{CNN-LSTM}, \textit{CNN-BiLSTM}, and \textit{CNN-BiGRU}. These models adopt a dual-stream structure, where both the convolutional and recurrent branches process the \textit{TF-IDF} representations independently before merging. The hyperparameters for this setting are 256, 128 units in the main layer, kernel size of 5, max-pooling with a size of 2, with a dropout rate of 0.3. The outputs from both streams were concatenated at last and passed through a unified classification head comprising fully connected layers with 128 and 64 units, activated by ReLU and regularized with dropout rates of 0.5 and 0.3, respectively.
\end{enumerate}
\noindent Training was performed using a stratified 3-fold cross-validation setup, with each fold trained for 3 epochs. We used the Adam optimizer with a batch size of 64, early stopping with a patience of 2, and learning rate reduction on plateau with sigmoid and softmax activations for binary and multiclass, respectively. 

\subsection{Transformer Based Classifiers}
We fine-tuned several transformer-based models for this classification task, including the base versions of Bidirectional Encoder Representations from Transformers (BERT) \cite{devlin2018bert}, DistilBERT \cite{sanh2019distilbert}, Robustly Optimized BERT Pretraining Approach (RoBERTa) \cite{liu2019roberta}, Decoding-enhanced BERT with disentangled attention (DeBERTa) \cite{he2020deberta}, and ModernBERT \cite{warner2024smarter}. The specific pretrained models used were \textit{bert-base-uncased, distilbert-base-uncased, roberta-base, deberta-v3-base}, and \textit{modernbert-base}. Two training configurations are explored for these models: (1) standard fine-tuning, where the transformer itself is trained end-to-end for the classification task, and (2) a Mixture-of-Experts (MoE) variant built on top of the transformer backbone. For the MoE configuration, we explored both \textit{Hard MoE}, where a gating mechanism selects a single expert per input, and \textit{Soft MoE}, where multiple experts contribute proportionally to the final prediction.

\begin{figure*}[htbp]
\centering
\includegraphics[width=0.5\linewidth]{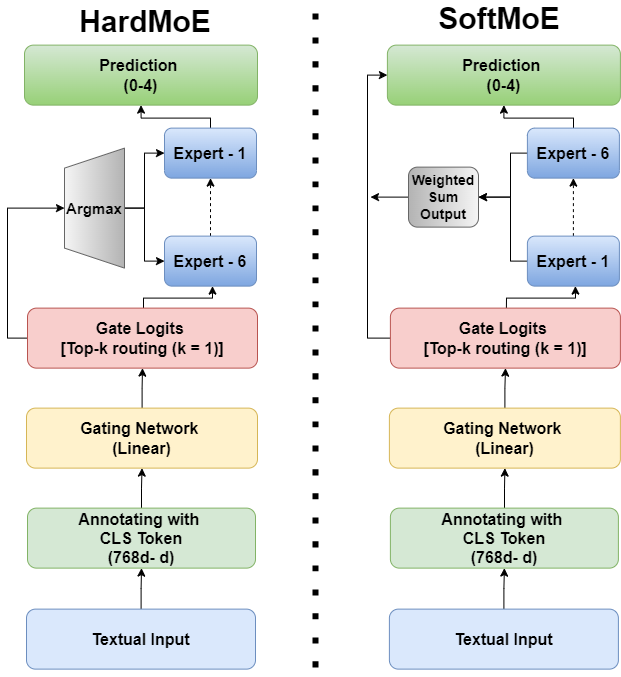} 
\caption{Pipeline of a novel Mixture-of-Experts (MoE) architecture integrated with Transformer models for AI text detection. The framework incorporates both Hard MoE (Left$\leftarrow$) and Soft MoE (Right$\rightarrow$) mechanisms to enhance detection performance across diverse input types.}\label{fig:hradsoftMoE}
\end{figure*}

\textbf{MoE Detector:} We employed two kinds of MoE architectures, namely \textit{\textbf{HardMoE}} and \textit{\textbf{SoftMoE}} classifiers. In the HardMoE Classifier, a hard gating mechanism is employed, which is a linear gating network that maps the CLS token (\texttt{Transformer(x)[:,0,:]}) to expert logits \texttt{$\mathbf{g}$ = $\mathbf{W}_g$ $\mathbf{h}_{\text{CLS}}$ + $\mathbf{b}_g$}, and the expert is selected with the highest logit computed through the argmax \texttt{$\arg\max_{i} (\mathbf{g}_i)$} operation. This selected expert will take the input, and the output of this chosen expert is passed through the softmax function for the prediction of class. Also, these gate logits are available for auxiliary loss computations. Unlike HardMoE, SoftMoE is driven by the soft gating method. The gating network here produces logits for all experts, and further applying the softmax, the final normalized weights are obtained. Such that, SoftMoE allows the model to consider all experts for each input. The actual difference between the two architectures is in the gating mechanisms, where HardMoE is like \textbf{Winner-Takes-All} selection, and SoftMoE is like aggregating outputs from all the Experts. The Algorithm \ref{hardsoftMoE} shows the Forward Pass functions, which tell how the input is changing with the layers and how the classification is done, and an illustration of the pipeline can be seen in the Fig~\ref{fig:hradsoftMoE}.

\begin{algorithm}[ht]
\begin{algorithmic}[1]
\caption{Forward Pass for MoE Classifier (Hard or Soft)}
\label{hardsoftMoE}
\State \textbf{Input:} $input\_ids$, $attention\_mask$
\State \textbf{Output:} $output\_logits$, $gate\_logits$
\State Extract $hidden\_state$ from base transformer
\State Get CLS token: $cls \leftarrow hidden\_state[:, 0, :]$
\State Apply dropout to $cls$
\State Compute: $gate\_logits \leftarrow \text{Linear}(cls)$
\If{model is HardMoE}
    \State $expert\_choice \leftarrow \arg\max(gate\_logits, \text{dim}=1)$
    \State Initialize $output\_logits$ as zeros
    \For{each expert $i$}
        \State $mask \leftarrow (expert\_choice == i)$
        \If{$mask$ not empty}
            \State $out \leftarrow expert_i(cls[mask])$
            \State $output\_logits[mask] \leftarrow out$
        \EndIf
    \EndFor
\Else { SoftMoE}
    \State $gate\_weights \leftarrow \text{Softmax}(gate\_logits)$
    \State $expert\_outputs \leftarrow$ []
    \For{each expert $i$}
        \State $out \leftarrow expert_i(cls)$
        \State Append $out$ to $expert\_outputs$
    \EndFor
    \State Stack $expert\_outputs$
    \State $output\_logits \leftarrow \sum (\text{gate\_weights} \times \text{expert\_outputs})$
\EndIf
\State \Return $output\_logits$, $gate\_logits$
\end{algorithmic}
\end{algorithm}
   
\subsection{Adversarial Addressing}
As shown in the figure above \ref{AllExperiments}, we address the Adversarial robustness in two different settings. 1) \textit{Adversarial Training}, 2) \textit{Adversarial Preprocessing and Implicit Adversarial Classification}. This is framed as a binary classification task; the goal is to accurately classify texts even under adversarial perturbations. To this end, we directly fine-tune standard transformer-based models for both settings, but have a slight difference.

\subsubsection{Adversarial Training}
This approach involves fine-tuning the models directly on the whole dataset containing both clean and adversarially perturbed samples. We fine-tune 5 transformer backbones: \textit{bert-base-uncased, distilbert-base-uncased, roberta-base, deberta-v3-base,} and \textit{modernbert-base}. 
Word embeddings are inherited from their respective transformers. Models are trained for 3 epochs with a batch size of 32, a maximum input length of 512 tokens, and early stopping with a patience of 2 epochs. Optimization is performed using the AdamW optimizer with a standard cross-entropy loss.

\subsubsection{Adversarial Preprocessing and Implicit Adversarial Classification}
We introduce this new method for addressing the classification task in the case of Adversarial attacks by the following pipeline: 
\begin{enumerate}
    \item The dataset consists of a \texttt{text} column (containing both clean and adversarial examples) and a corresponding \texttt{label} (human or AI).
    \item We preprocess the texts in \texttt{text} column to generate a \texttt{preprocessed\_text} column, aiming to minimize adversarial artifacts.
    \item We compute a set of features comparing the original \texttt{text} column and \texttt{preprocessed text} column: Cosine Similarity, Edit Distance, Word-Overlap Ratio, Homoglyph Substitution Count, BLEU Score, and Word Error Rate (WER).
    \item  These computed features, along with the original (potentially attacked) text, are input into the model for classification.
    \item Notably, we do not explicitly indicate whether a sample is adversarial or not; the model implicitly learns to differentiate based on these comparative features.
\end{enumerate}

\noindent For example, in terms of Cosine-Similarity, if a text $x$ is unperturbed, its preprocessed version $x'$, which remains identical ($x' = x$), results in a cosine similarity of 1. In contrast, adversarial modifications yield $x' \neq x$, and the similarity drops below 1. This difference serves as a soft signal for implicit adversarial detection. As shown in Table~\ref{tab:adversarial_binary_classification}, the implicit detection model outperforms the adversarially trained model across both classes.

\subsection{Sentence Segmentation} 
This section outlines the approach for the sentence segmentation task in mixed collaborative texts composed of both human-written and AI-generated spans. As detailed in the Dataset descriptions section, the dataset comprises three categories: 1) H$\rightarrow$M, 2) M$\rightarrow$H, and 3) Completely Human-AI Mixed text. The complete dataset has around 50,000 samples: 20k from each Type 1 and 2, and 10k from Type 3. To solve the boundary detection task, we implemented experiments using Conditional Random Fields (CRFs), coupled with different Transformer and Neural Network backbones. We tested three architectural configurations: 1) NN + CRF, 2) Transformer + CRF, and 3) Transformer + NN + CRF.

\begin{enumerate}
    \item \textbf{NN + CRF (Neural Network + CRF):} This is a hybrid structure that comprises in a hierarchical way with NN backbone, and a CRF layer. We used the \texttt{deberta-v3-base} model to produce contextual token embeddings. These embeddings were subsequently fed into one of five different neural architectures: CNN, RNN, LSTM, BiLSTM, and BiGRU. A dropout layer and a linear projection were then used prior to feeding into the CRF layer that encodes consistent label transitions. This produced five different experiments.

    \item \textbf{Transformer + CRF:} We substituted the neural backbone with different transformer encoders in this setup. The CRF layer was added on top of the last transformer representations to capture sequential tagging. Layer-wise learning rate dropping, dynamic dropout, and Xavier initialization were used to enhance generalization and training stability. We tried five transformer models: BERT, DistilBERT, RoBERTa, DeBERTa, and ModernBERT, which gave us five Transformer-CRF experiments.

    \item \textbf{Transformer + NN + CRF:} This hybrid system integrates the advantages of both transformer and neural network backbones. Two experimental approaches were adopted: (a) Best Transformer + All NNs + CRF, (b) All Transformers + Best NN + CRF. DeBERTa was chosen as the best transformer backbone, and BiGRU as the best NN backbone based on prior experiment performance. In (a), DeBERTa was coupled with every one of the five NNs (CNN, RNN, LSTM, BiLSTM, BiGRU). In (b), BiGRU was combined with each of the five transformers. Since both strategies share the same DeBERTa-BiGRU-CRF model, we removed the duplicates and had nine different experiments under this hybrid setup.
\end{enumerate}

\section{Experimental Setup} \label{sec:experimentalsetup}
We conducted all our experiments on Amazon Web Services (\textit{AWS}) Cloud server, Amazon Elastic Compute Cloud (\textit{EC2}) instance. In the EC2 instance, we initiated an instance for Accelerated Computing. The specifications are \textbf{g6e.xlarge} instance, which provides 3rd generation AMD EPYC processors (\textit{AMD EPYC 7R13}), with a \textbf{NVIDIA L40S} Tensor Core GPU with 48 GB GPU memory, and 4x vCPU with 32 GiB memory and a network bandwidth of 20GBps, and our OS type is Ubuntu Server 24.04 LTS (\textit{HVM}).

\section{Results and Discussions}\label{sec:res_discussion}
As we have done a vast set of experiments, the concise and best results are shown for different tasks at different Tables \ref{tab:binary_results}, \ref{tab:adversarial_binary_classification}, \ref{tab:multi_label_models}, \ref{tab:all_multiclass_classification_clean}. 
Table \ref{tab:binary_results} gives all set of models over Binary Classification. Table \ref{tab:adversarial_binary_classification} has the results of model setting over the Adversarial Data. The Table \ref{tab:multi_label_models} gives the results of Sentence Segmentation models. Finally, Table \ref{tab:all_multiclass_classification_clean} shows the performance of all models over Multiclass Classification. Binary Classification task is benchmarked with both Recall and F1-score, with the highest values to Recall and F1-Score of 99.87\% and 99.45\% for the Human-Class, while 99.02\% and 99.44\% for AI-Class. Multiclass Classification, ModernBERT-Transformer has the Highest performing Accuracy of 94.57\%, outperforming newly proposed HardMoE and SoftMoE Detector architectures. For the Sentence Segmentation task, the benchmark Evaluation metric is Matthews Correlation Coefficient (\textit{MCC}), while the model \textit{DeBERTa-BiGRU-CRF} got the highest score of 97.89\%. Among the two methodologies that are discussed for the Adversarial Detection and binary classification correctly, the newly proposed method got the highest values of Recall and F1-score of classes Human as 93.63\%, 98.22\% and class AI as 94.77\% and 94.24\%, respectively.

\begin{table}[htbp]
\centering
\caption{Performance of various models across feature representations for \textbf{binary classification}.}
\label{tab:binary_results}
\scriptsize
\begin{tabular}{ccccccc}
\toprule
\rotatebox{90}{\textbf{}} & \textbf{Model} & \multicolumn{2}{c}{\textbf{Human}} & \multicolumn{2}{c}{\textbf{AI}} & \textbf{Accuracy} \\
\cmidrule(lr){3-4} \cmidrule(lr){5-6}
& & \textbf{Recall} & \textbf{F1} & \textbf{Recall} & \textbf{F1} & \\
\midrule
\multirow{5}{*}{\rotatebox{90}{NELA}} 
& LR & 47.16 & 50.03 & 60.32 & 56.77 & 53.78 \\
& RF & \textbf{61.35} & 62.95 & 66.83 & 65.20 & 64.11 \\
& XGB & 57.73 & 64.32 & \textbf{78.48} & \textbf{71.27} & 68.17 \\
& LDA & 45.78 & 49.70 & 62.02 & 57.55 & 53.96 \\
& SVM & 56.95 & \textbf{65.44} & 74.63 & 63.00 & 58.91 \\
\cmidrule(lr){1-7}
\multirow{5}{*}{\rotatebox{90}{Word2Vec}} 
& LR & 96.15 & 96.56 & 97.04 & 96.63 & 96.60 \\
& RF & 96.45 & 96.10 & 95.57 & 96.12 & 96.11 \\
& XGB & \textbf{97.66} & \textbf{97.64} & \textbf{97.66} & \textbf{97.67} & \textbf{97.66} \\
& LDA & 96.65 & 96.08 & 95.52 & 96.08 & 96.08 \\
& SVM & 96.02 & 96.52 & 97.09 & 96.60 & 96.56 \\
\cmidrule(lr){1-7}
\multirow{5}{*}{\rotatebox{90}{TF-IDF}} 
& LR & 98.21 & 97.94 & 97.68 & 97.95 & 97.95 \\
& RF & 98.21 & 97.90 & 97.61 & 97.91 & 97.91 \\
& XGB & \textbf{98.64} & \textbf{98.33} & \textbf{98.03} & \textbf{98.34} & \textbf{98.33} \\
& LDA & 98.33 & 98.03 & 97.73 & 98.04 & 98.03 \\
& SVM & 98.33 & 98.17 & 98.03 & 98.19 & 98.18 \\
\cmidrule(lr){1-7}
\multirow{5}{*}{\rotatebox{90}{No Hybrid}} 
& CNN & 91.54 & 95.33 & \textbf{99.49} & 95.67 & 95.51 \\
& RNN & \textbf{98.62} & \textbf{97.85} & 97.04 & \textbf{97.81} & \textbf{97.83} \\
& LSTM & 97.75 & 97.77 & 97.79 & 97.77 & 97.77 \\
& BiLSTM & 98.32 & 97.77 & 97.19 & 97.74 & 97.76 \\
& BiGRU & 97.18 & 97.61 & 98.07 & 97.63 & 97.62 \\
\cmidrule(lr){1-7}
\multirow{3}{*}{\rotatebox{90}{Hybrid}} 
& CNN-LSTM & 99.12 & 98.00 & 96.81 & 97.94 & 97.97 \\
& CNN-BiLSTM & 98.42 & \textbf{98.42} & \textbf{98.42} & \textbf{98.42} & \textbf{98.42} \\
& CNN-BiGRU & \textbf{99.42} & 97.87 & 96.24 & 97.79 & 97.83 \\
\cmidrule(lr){1-7}
\multirow{5}{*}{\rotatebox{90}{Normal}} 
& BERT & 96.18 & 98.00 & 99.89 & 98.06 & 98.03 \\
& DistilBERT & 98.75 & 99.24 & 99.74 & 99.25 & 99.25 \\
& RoBERTa & 94.16 & 96.96 & 99.94 & 97.12 & 97.05 \\
& DeBERTa & 96.78 & 98.35 & \textbf{99.97} & 98.39 & 98.37 \\
& ModernBERT & \textbf{99.87} & \textbf{99.45} & 99.02 & \textbf{99.44} & \textbf{99.45} \\
\bottomrule
\end{tabular}
\end{table}

\begin{table}[htbp]
\centering
\caption{Performance comparison of transformer-based models under \textbf{Adversarial Training} and \textbf{Implicit Method} settings for binary classification of human-written vs.\ AI-generated texts. Results are reported for recall, F1-score, and overall accuracy.}
\label{tab:adversarial_binary_classification}
\scriptsize
\begin{tabular}{ccccccc}
\toprule
\rotatebox{90}{\textbf{}} & \multirow{2}{*}{\textbf{Model}} 
& \multicolumn{2}{c}{\textbf{Human}} & \multicolumn{2}{c}{\textbf{AI}} & \multirow{2}{*}{\textbf{Accuracy}} \\
\cmidrule(lr){3-4} \cmidrule(lr){5-6}
& & \textbf{Recall} & \textbf{F1} & \textbf{Recall} & \textbf{F1} & \\
\midrule
\multirow{5}{*}{\rotatebox{90}{\shortstack{Adversarial\\Training}}} 
    & BERT        & 86.21 & 88.93 & 89.39 & 88.23 & 88.12 \\
    & DistilBERT  & 88.57 & 89.04 & 89.85 & \textbf{89.87} & 89.00 \\
    & RoBERTa     & 84.96 & 86.24 & 89.14 & 87.45 & 87.47 \\
    & DeBERTa     & 86.98 & 88.55 & \textbf{89.99} & 88.49 & 88.90 \\
    & ModernBERT  & \textbf{89.03} & \textbf{89.97} & 89.46 & 89.64 & \textbf{89.55} \\
\cmidrule(lr){1-7}
\multirow{5}{*}{\rotatebox{90}{\shortstack{Implicit\\Method}}} 
    & BERT        & \textbf{94.34} & 94.44 & 95.57 & \textbf{94.45} & \textbf{94.46} \\
    & DistilBERT  & 92.99 & 94.24 & 95.54 & 94.28 & 94.26 \\
    & RoBERTa     & 92.02 & 93.59 & 95.24 & 93.66 & 93.63 \\
    & DeBERTa     & 92.78 & 94.21 & \textbf{95.69} & 94.25 & 94.23 \\
    & ModernBERT  & 93.69 & \textbf{98.22} & 94.77 & 94.24 & 94.23 \\
\bottomrule
\end{tabular}
\end{table}

\begin{table}[htbp]
\centering
\caption{Performance of neural and transformer-based CRF models for multi-label classification, showing accuracy, precision, recall, F1-score, and MCC.}
\label{tab:multi_label_models}
\scriptsize
\begin{tabular}{c c c c c c c}
\toprule
\rotatebox{90}{\textbf{}} & \textbf{Model} & \textbf{Accuracy} & \textbf{Precision} & \textbf{Recall} & \textbf{F1-score} & \textbf{MCC} \\
\midrule
\multirow{5}{*}{\rotatebox{90}{\shortstack{NN\\CRF}}} 
    & CNN CRF & 90.89 & 90.90 & 90.89 & 90.86 & 81.34 \\
    & RNN CRF & 90.03 & 90.02 & 90.03 & 90.00 & 79.56 \\
    & LSTM CRF & 93.74 & 93.76 & 93.74 & 93.73 & 87.19 \\
    & BiLSTM CRF & 95.58 & 95.58 & 95.58 & 95.57 & 90.95 \\
    & \textbf{BiGRU CRF} & \textbf{96.19} & \textbf{96.19} & \textbf{96.19} & \textbf{96.19} & \textbf{92.21} \\
\midrule
\multirow{5}{*}{\rotatebox{90}{\shortstack{Transfor-\\mer CRF}}} 
    & BERT CRF & 97.89 & 97.90 & 97.89 & 97.89 & 95.76 \\
    & DistilBERT CRF & 97.52 & 97.54 & 97.52 & 97.52 & 95.03 \\
    & RoBERTa CRF & 98.41 & 98.42 & 98.41 & 98.41 & 96.81 \\
    & ModernBERT CRF & 98.72 & 98.72 & 98.72 & 98.72 & 97.41 \\
    & \textbf{DeBERTa CRF} & \textbf{98.84} & \textbf{98.84} & \textbf{98.84} & \textbf{98.84} & \textbf{97.67} \\
\midrule
\multirow{10}{*}{\rotatebox{90}{\shortstack{Transformer\\NN CRF}}} 
    & BERT BiGRU CRF & 97.96 & 97.98 & 97.96 & 97.96 & 95.92 \\
    & DeBERTa CNN CRF & 98.83 & 98.83 & 98.83 & 98.83 & 97.64 \\
    & DeBERTa RNN CRF & 98.81 & 98.81 & 98.81 & 98.81 & 97.61 \\
    & DeBERTa LSTM CRF & 98.86 & 98.86 & 98.86 & 98.86 & 97.69 \\
    & RoBERTa BiGRU CRF & 98.48 & 98.50 & 98.48 & 98.48 & 96.96 \\
    & DeBERTa BiLSTM CRF & 98.88 & 98.88 & 98.88 & 98.88 & 97.74 \\
    & \textbf{DeBERTa BiGRU CRF} & \textbf{98.95} & \textbf{98.95} & \textbf{98.95} & \textbf{98.95} & \textbf{97.89} \\
    & DistilBERT BiGRU CRF & 97.69 & 97.70 & 97.69 & 97.69 & 95.35 \\
    & ModernBERT BiGRU CRF & 98.70 & 98.70 & 98.70 & 98.70 & 97.37 \\
\bottomrule
\end{tabular}
\end{table}

\begin{table*}[htbp]
\centering
\caption{Comprehensive MultiClass classification results across various models and AI text sources.}
\label{tab:all_multiclass_classification_clean}
\scriptsize
\resizebox{1.0\linewidth}{!}{%
\begin{tabular}{c c cc cc cc cc cc c}
\toprule
\multirow{2}{*}{\rotatebox{90}{\textbf{}}} & \multirow{2}{*}{\textbf{Model}} & \multicolumn{2}{c}{\textbf{Human}} & \multicolumn{2}{c}{\textbf{OpenAI}} & \multicolumn{2}{c}{\textbf{Anthropic}} & \multicolumn{2}{c}{\textbf{Deepseek}} & \multicolumn{2}{c}{\textbf{Llama}} & \multirow{2}{*}{\textbf{Accuracy}} \\
\cmidrule(lr){3-4} \cmidrule(lr){5-6} \cmidrule(lr){7-8} \cmidrule(lr){9-10} \cmidrule(lr){11-12} 
& & \textbf{Recall} & \textbf{F1} & \textbf{Recall} & \textbf{F1} & \textbf{Recall} & \textbf{F1} & \textbf{Recall} & \textbf{F1} & \textbf{Recall} & \textbf{F1} & \\
\midrule
\multirow{5}{*}{\rotatebox{90}{Word2Vec}} & LR & 96.62 & 95.85 & 64.58 & 63.32 & 63.14 & 62.03 & 54.81 & 57.70 & 59.16 & 59.42 & 78.42 \\
& RF & 98.89 & 93.85 & 57.73 & 57.25 & 51.79 & 55.17 & 52.27 & 53.63 & 42.69 & 49.18 & 74.83 \\
& XGB & 98.33 & 96.91 & 62.50 & 64.64 & 68.22 & 66.79 & 61.49 & 62.09 & 60.42 & 62.68 & 80.63 \\
& LDA & 97.18 & 95.58 & 63.78 & 63.38 & 60.15 & 60.49 & 53.59 & 55.36 & 56.14 & 57.68 & 77.68 \\
& SVM & 97.23 & 96.04 & 64.78 & 65.13 & 62.94 & 62.41 & 55.62 & 58.71 & 60.33 & 60.33 & 78.97 \\
\midrule
\multirow{5}{*}{\rotatebox{90}{TF-IDF}} & LR & 99.09 & 97.17 & 84.72 & 85.32 & 85.05 & 85.78 & 80.64 & 82.91 & 81.28 & 84.03 & 90.96 \\
& RF & 99.94 & 92.32 & 72.02 & 75.38 & 74.50 & 81.43 & 65.65 & 71.01 & 65.59 & 74.57 & 84.60 \\
& XGB & 99.49 & 97.84 & 82.73 & 84.45 & 86.45 & 87.32 & 82.16 & 82.58 & 83.52 & 86.17 & 91.56 \\
& LDA & 98.64 & 97.56 & 82.04 & 83.40 & 84.36 & 86.64 & 82.06 & 81.57 & 85.08 & 85.67 & 90.97 \\
& SVM & 99.04 & 97.76 & 85.61 & 85.87 & 86.45 & 85.85 & 82.06 & 83.37 & 84.32 & 84.32 & 91.36 \\
\midrule
\multirow{5}{*}{\rotatebox{90}{No Hybrid}} & CNN & 99.97 & 86.19 & 71.35 & 69.41 & 57.81 & 67.15 & 99.00 & 32.97 & 57.41 & 65.66 & 75.73 \\
& RNN & 98.60 & 97.29 & 83.24 & 82.14 & 82.32 & 81.88 & 78.01 & 78.76 & 77.19 & 80.99 & 89.21 \\
& LSTM & 99.67 & 97.35 & 81.90 & 82.75 & 80.17 & 82.42 & 77.82 & 80.45 & 82.74 & 81.29 & 89.65 \\
& BiLSTM & 97.43 & 97.41 & 86.97 & 82.89 & 81.93 & 83.35 & 77.62 & 80.08 & 81.23 & 81.19 & 89.67 \\
& BiGRU & 98.60 & 97.47 & 86.76 & 81.41 & 79.88 & 83.17 & 73.96 & 79.00 & 80.82 & 81.23 & 89.46 \\
\midrule
\multirow{3}{*}{\rotatebox{90}{Hybrid}} & CNN-LSTM & 99.37 & 97.70 & 86.14 & 83.21 & 85.64 & 84.61 & 75.04 & 80.89 & 80.22 & 83.77 & 90.57 \\
& CNN-BiLSTM & 99.87 & 96.12 & 81.59 & 82.61 & 83.10 & 84.67 & 76.93 & 80.04 & 74.97 & 80.09 & 89.53 \\
& CNN-BiGRU & 98.97 & 98.26 & 84.17 & 83.44 & 83.39 & 85.27 & 82.57 & 83.10 & 83.35 & 84.07 & 91.18 \\
\midrule
\multirow{4}{*}{\rotatebox{90}{\shortstack{Hard\\MoE}}} & BERT & 96.28 & 98.00 & 80.97 & 82.63 & 88.37 & 86.81 & 71.68 & 79.42 & 93.54 & 79.84 & 89.97 \\
& DistilBERT & 93.86 & 96.69 & 76.52 & 81.49 & 80.37 & 85.24 & 79.00 & 78.38 & 95.96 & 77.91 & 88.42 \\
& RoBERTa & 97.30 & 98.47 & 78.90 & 77.46 & 87.50 & 84.72 & 77.03 & 76.18 & 90.61 & 80.07 & 89.05 \\
& DeBERTa & 93.83 & 96.77 & 75.07 & 79.73 & 95.99 & 77.31 & 73.36 & 78.41 & 80.92 & 81.29 & 87.66 \\
\midrule
\multirow{4}{*}{\rotatebox{90}{\shortstack{Soft\\MoE}}} & BERT & 95.23 & 97.45 & 83.14 & 81.41 & 84.96 & 86.65 & 72.47 & 77.21 & 95.15 & 82.43 & 89.57 \\
& DistilBERT & 93.98 & 96.82 & 80.97 & 81.81 & 87.69 & 86.59 & 80.29 & 78.96 & 93.64 & 85.25 & 89.83 \\
& RoBERTa & 95.50 & 97.61 & 81.69 & 80.61 & 77.73 & 84.81 & 84.55 & 75.84 & 87.08 & 83.46 & 89.13 \\
& DeBERTa & 97.72 & 98.68 & 84.59 & 84.59 & 97.26 & 83.24 & 78.71 & 82.38 & 77.90 & 85.34 & 91.22 \\
\midrule
\multirow{5}{*}{\rotatebox{90}{Normal}} & BERT & 97.85 & 98.86 & 96.03 & 85.86 & 82.42 & 89.17 & 84.55 & 84.72 & 96.56 & 86.25 & 92.61 \\
& DistilBERT & 98.35 & 99.03 & 78.07 & 83.98 & 95.01 & 88.33 & 83.26 & 84.86 & 90.11 & 86.86 & 92.55 \\
& RoBERTa & 96.48 & 98.84 & 78.80 & 84.38 & 89.94 & 89.37 & 83.86 & 84.78 & 95.66 & 84.07 & 91.81 \\
& DeBERTa & 97.33 & 98.63 & 80.45 & 85.96 & 92.18 & 89.73 & 82.57 & 85.23 & 96.46 & 86.16 & 92.66 \\
& ModernBERT & 99.52 & 99.57 & 93.48 & 87.89 & 88.28 & 92.29 & 85.64 & 86.62 & 91.22 & 91.54 & 94.57 \\
\bottomrule
\end{tabular}
}
\end{table*}

\section{Ablation Study}
\label{sec:ablation}
\subsection{Comparison with Zero-shot Methods}

Tables~\ref{tab:zero_shot_binary} and~\ref{tab:zero_shot_adv} compare our ModernBERT model with representative zero-shot AI text detectors (\textit{Binoculars}, \textit{Glimpse}, and \textit{FastDetect}). \textbf{Binary Classification:} From Table~\ref{tab:zero_shot_binary}, zero-shot baselines achieve moderate performance (accuracy $72.20\%$–$96.40\%$), whereas ModernBERT attains near-perfect results (\textbf{99.87\%} Human recall, \textbf{99.02\%} AI recall, \textbf{99.45\%} accuracy), demonstrating strong generalization beyond large zero-shot LLMs. \textbf{Adversarial Binary Classification:} In Table~\ref{tab:zero_shot_adv}, zero-shot methods collapse under adversarial inputs (AI recall $0.65\%$–$27\%$, accuracy $<37\%$).  
ModernBERT remains robust (\textbf{94\%+} recall/F1, \textbf{94.46\%} accuracy), underscoring the benefits of task-specific fine-tuning.

\begin{table}[htbp]
\centering
\caption{Performance comparison with Zero-shot methods for \textbf{binary classification}.}
\label{tab:zero_shot_binary}
\begin{tabular}{ccccccc}
\toprule
\rotatebox{90}{\textbf{}} & \textbf{Model} & \multicolumn{2}{c}{\textbf{Human}} & \multicolumn{2}{c}{\textbf{AI}} & \textbf{Accuracy} \\
\cmidrule(lr){3-4} \cmidrule(lr){5-6}
& & \textbf{Recall} & \textbf{F1} & \textbf{Recall} & \textbf{F1} & \\
\midrule
\multirow{5}{*}{\rotatebox{90}{Zero-shot}} 
& Binoculars & 90.00 & 56.42 & 67.75 & 79.58 & 72.20 \\
& Glimpse (davinci) & 91.00 & 63.01 & 71.49 & 83.12 & 76.81 \\
& Glimpse (babbage) & 92.00 & 70.93 & 77.04 & 87.58 & 80.23 \\
& FastDetect(falcon-7b) & 92.00 & 91.08 & 97.50 & 97.74 & 96.40 \\
& FastDetect(gpt-neo-2.7b) & 92.00 & 72.15 & 84.25 & 90.46 & 85.80 \\
\cmidrule(lr){1-7}
& ModernBERT (Best) & \textbf{99.87} & \textbf{99.45} & \textbf{99.02} & \textbf{99.44} & \textbf{99.45} \\
\bottomrule
\end{tabular}
\end{table}

\begin{table}[htbp]
\centering
\caption{Performance comparison with Zero-shot methods for \textbf{binary classification in Adversarial Scenario}.}
\label{tab:zero_shot_adv}
\begin{tabular}{ccccccc}
\toprule
\rotatebox{90}{\textbf{}} & \multirow{2}{*}{\textbf{Model}} 
& \multicolumn{2}{c}{\textbf{Human}} & \multicolumn{2}{c}{\textbf{AI}} & \multirow{2}{*}{\textbf{Accuracy}} \\
\cmidrule(lr){3-4} \cmidrule(lr){5-6}
& & \textbf{Recall} & \textbf{F1} & \textbf{Recall} & \textbf{F1} & \\
\midrule
\multirow{5}{*}{\rotatebox{90}{Zero-shot}} 
    & Binoculars & 100.00 & 33.47 & 0.65 & 1.29 & 20.52 \\
    & Glimpse (davinci) & 84.60 & 31.24 & 10.75 & 18.76 & 25.63 \\
    & Glimpse (babbage) & 77.40 & 29.82 & 14.60 & 24.28 & 27.16 \\
    & FastDetect(falcon-7b) & 89.60 & 35.51 & 21.25 & 34.31 & 34.92 \\
    & FastDetect(gpt-neo-2.7b) &75.20 & 32.16 & 26.90 & 40.42 & 36.56 \\
\cmidrule(lr){1-7}
& BERT (Best)      & \textbf{94.34} & \textbf{94.44} & \textbf{95.57} & \textbf{94.45} & \textbf{94.46} \\
\bottomrule
\end{tabular}
\end{table}

\subsection{Ablation on Number of Experts in MoE}

Table~\ref{tab:ablation_moe} summarizes the impact of varying the number of experts in HardMoE (BERT backbone) and SoftMoE (DeBERTa backbone) for multi-class classification. In both variants, the \textbf{1-expert} configuration achieves the highest accuracy (92.61\% for HardMoE, 92.66\% for SoftMoE), outperforming 3- and 6-expert setups. Adding more experts leads to inconsistent performance across source categories, with particularly large recall drops for Deepseek and Llama, despite strong results on Human text. These results suggest that while increasing experts expands capacity, it may reduce specialization and weaken generalization across diverse generative models.

\begin{table*}[htbp]
\centering
\caption{Ablation Study: Expert Count in MoE for the Multi-class classification for the models that performed best in HardMoE and SoftMoE Variations, where BERT in HardMoE and DeBERTa in SoftMoE.}
\label{tab:ablation_moe}
\scriptsize
\resizebox{1.0\linewidth}{!}{%
\begin{tabular}{c c cc cc cc cc cc c}
\toprule
\multirow{2}{*}{\rotatebox{90}{\textbf{}}} & \multirow{2}{*}{\textbf{Model}} & \multicolumn{2}{c}{\textbf{Human}} & \multicolumn{2}{c}{\textbf{OpenAI}} & \multicolumn{2}{c}{\textbf{Anthropic}} & \multicolumn{2}{c}{\textbf{Deepseek}} & \multicolumn{2}{c}{\textbf{Llama}} & \multirow{2}{*}{\textbf{Accuracy}} \\
\cmidrule(lr){3-4} \cmidrule(lr){5-6} \cmidrule(lr){7-8} \cmidrule(lr){9-10} \cmidrule(lr){11-12}
& & \textbf{Recall} & \textbf{F1} & \textbf{Recall} & \textbf{F1} & \textbf{Recall} & \textbf{F1} & \textbf{Recall} & \textbf{F1} & \textbf{Recall} & \textbf{F1} & \\
\midrule
\multirow{3}{*}{\rotatebox{90}{\shortstack{Hard\\MoE}}} & BERT (1-expert) & 97.85 & 98.86 & 96.03 & 85.86 & 82.42 & 89.17 & 84.55 & 84.72 & 96.56 & 86.25 & 92.61 \\
& BERT (3-experts) & 97.21 & 98.43 & 88.51 & 84.38 & 86.91 & 87.34 & 76.23 & 81.66 & 94.85 & 82.11 & 91.04 \\
& BERT (6-experts) & 96.28 & 98.00 & 80.97 & 82.63 & 88.37 & 86.81 & 71.68 & 79.42 & 93.54 & 79.84 & 89.97 \\
\midrule
\multirow{3}{*}{\rotatebox{90}{\shortstack{Soft\\MoE}}} & DeBERTa (1-expert) & 97.33 & 98.63 & 80.45 & 85.96 & 92.18 & 89.73 & 82.57 & 85.23 & 96.46 & 86.16 & 92.66 \\
& DeBERTa (3-experts) & 97.52 & 98.66 & 82.66 & 85.24 & 95.42 & 85.91 & 80.31 & 84.22 & 87.65 & 85.72 & 91.74 \\
& DeBERTa (6-experts) & 97.72 & 98.68 & 84.59 & 84.59 & 97.26 & 83.24 & 78.71 & 82.38 & 77.90 & 85.54 & 91.22 \\
\bottomrule
\end{tabular}
}
\end{table*}

\section{Conclusion and Future Scope} \label{sec:conclusion}
In this paper, we have presented a vast set of experiments regarding the AI text detection. The experiments are done from the very basic to the most recent and advanced approaches. All these current methods comes under Supervised Learning, we haven't explored the Zero-shot or One-shot methods. So, we are expected to do these experiments in the future self. We found that, the Detectors are performing well in binary classification on, but they are getting more confused when the case of Multi-Class and Bi-Label comes into the picture. 
Overall the detectors, DeBERTa and ModernBERT transformer models are giving the best results on average. 

As this paper only focused on \textit{``Text''}, we wanted to explore other AI generated content like Images and Speech, and do experimentation on these variations. For the case of Adversarial Attacks, we have explored one method, and proposed one new method, so it is worth to lookup other methods.

\section{Limitations} \label{sec:limitations}
In this section, we discuss some of the key limitations of our work. We mainly focused on the English language, but neither focused on multi- nor low-resource languages. Even though english is the major language used worldwide, but there are cases where their own language is used in Academics, etc. We have taken a very few set of LLM's, which includes two open source and two closed source models, but the inclusion of more open source models can also make the dataset much more robust. But even though there are few models, we have taken the latest versions of all at the time of dataset creation. All of our experiments are only in one case: supervised training and fine-tuning. These methods, especially in the case of Transformer models are included fine-tuning consume a lot of time. 

\section{Ethics}
We hereby declare that this E-BMAS dataset was created by only using the existing publicly available dataset for the Human corpus and has not been taken or scraped from any website or any online platform. All the datasets taken are properly cited in this paper. For the AI texts, we have taken the API keys for each AI model and sent API requests for the Chat-Completions.
\section*{Acknowledgements}
We are grateful to the Department of Computer Science and Engineering at the National Institute of Technology, Silchar for allowing us to conduct our research and experiments. We also extend our thanks to the CNLP \& AI laboratories for their valuable resources and the conducive research environment.

\section*{Declarations}
\begin{itemize}
\item Funding: This research received no external funding.
 \item Conflict of interest: The authors declare no conflict of interest.
 \item Data availability: Data is available on the Huggingface given above.

\end{itemize}

\backmatter
\bibliography{sn-bibliography}


\begin{thebibliography}{44}
\ifx \bisbn   \undefined \def \bisbn  #1{ISBN #1}\fi
\ifx \binits  \undefined \def \binits#1{#1}\fi
\ifx \bauthor  \undefined \def \bauthor#1{#1}\fi
\ifx \batitle  \undefined \def \batitle#1{#1}\fi
\ifx \bjtitle  \undefined \def \bjtitle#1{#1}\fi
\ifx \bvolume  \undefined \def \bvolume#1{\textbf{#1}}\fi
\ifx \byear  \undefined \def \byear#1{#1}\fi
\ifx \bissue  \undefined \def \bissue#1{#1}\fi
\ifx \bfpage  \undefined \def \bfpage#1{#1}\fi
\ifx \blpage  \undefined \def \blpage #1{#1}\fi
\ifx \burl  \undefined \def \burl#1{\textsf{#1}}\fi
\ifx \doiurl  \undefined \def \doiurl#1{\url{https://doi.org/#1}}\fi
\ifx \betal  \undefined \def \betal{\textit{et al.}}\fi
\ifx \binstitute  \undefined \def \binstitute#1{#1}\fi
\ifx \binstitutionaled  \undefined \def \binstitutionaled#1{#1}\fi
\ifx \bctitle  \undefined \def \bctitle#1{#1}\fi
\ifx \beditor  \undefined \def \beditor#1{#1}\fi
\ifx \bpublisher  \undefined \def \bpublisher#1{#1}\fi
\ifx \bbtitle  \undefined \def \bbtitle#1{#1}\fi
\ifx \bedition  \undefined \def \bedition#1{#1}\fi
\ifx \bseriesno  \undefined \def \bseriesno#1{#1}\fi
\ifx \blocation  \undefined \def \blocation#1{#1}\fi
\ifx \bsertitle  \undefined \def \bsertitle#1{#1}\fi
\ifx \bsnm \undefined \def \bsnm#1{#1}\fi
\ifx \bsuffix \undefined \def \bsuffix#1{#1}\fi
\ifx \bparticle \undefined \def \bparticle#1{#1}\fi
\ifx \barticle \undefined \def \barticle#1{#1}\fi
\bibcommenthead
\ifx \bconfdate \undefined \def \bconfdate #1{#1}\fi
\ifx \botherref \undefined \def \botherref #1{#1}\fi
\ifx \url \undefined \def \url#1{\textsf{#1}}\fi
\ifx \bchapter \undefined \def \bchapter#1{#1}\fi
\ifx \bbook \undefined \def \bbook#1{#1}\fi
\ifx \bcomment \undefined \def \bcomment#1{#1}\fi
\ifx \oauthor \undefined \def \oauthor#1{#1}\fi
\ifx \citeauthoryear \undefined \def \citeauthoryear#1{#1}\fi
\ifx \endbibitem  \undefined \def \endbibitem {}\fi
\ifx \bconflocation  \undefined \def \bconflocation#1{#1}\fi
\ifx \arxivurl  \undefined \def \arxivurl#1{\textsf{#1}}\fi
\csname PreBibitemsHook\endcsname

\bibitem[\protect\citeauthoryear{Chang et~al.}{2024}]{chang2024survey}
\begin{barticle}
\bauthor{\bsnm{Chang}, \binits{Y.}},
\bauthor{\bsnm{Wang}, \binits{X.}},
\bauthor{\bsnm{Wang}, \binits{J.}},
\bauthor{\bsnm{Wu}, \binits{Y.}},
\bauthor{\bsnm{Yang}, \binits{L.}},
\bauthor{\bsnm{Zhu}, \binits{K.}},
\bauthor{\bsnm{Chen}, \binits{H.}},
\bauthor{\bsnm{Yi}, \binits{X.}},
\bauthor{\bsnm{Wang}, \binits{C.}},
\bauthor{\bsnm{Wang}, \binits{Y.}}, \betal:
\batitle{A survey on evaluation of large language models}.
\bjtitle{ACM transactions on intelligent systems and technology}
\bvolume{15}(\bissue{3}),
\bfpage{1}--\blpage{45}
(\byear{2024})
\end{barticle}
\endbibitem

\bibitem[\protect\citeauthoryear{Annepaka and Pakray}{2025}]{annepaka2025large}
\begin{barticle}
\bauthor{\bsnm{Annepaka}, \binits{Y.}},
\bauthor{\bsnm{Pakray}, \binits{P.}}:
\batitle{Large language models: a survey of their development, capabilities, and applications}.
\bjtitle{Knowledge and Information Systems}
\bvolume{67}(\bissue{3}),
\bfpage{2967}--\blpage{3022}
(\byear{2025})
\end{barticle}
\endbibitem

\bibitem[\protect\citeauthoryear{Vaswani et~al.}{2017}]{vaswani2017attention}
\begin{botherref}
\oauthor{\bsnm{Vaswani}, \binits{A.}},
\oauthor{\bsnm{Shazeer}, \binits{N.}},
\oauthor{\bsnm{Parmar}, \binits{N.}},
\oauthor{\bsnm{Uszkoreit}, \binits{J.}},
\oauthor{\bsnm{Jones}, \binits{L.}},
\oauthor{\bsnm{Gomez}, \binits{A.N.}},
\oauthor{\bsnm{Kaiser}, \binits{{\L}.}},
\oauthor{\bsnm{Polosukhin}, \binits{I.}}:
Attention is all you need.
Advances in neural information processing systems
\textbf{30}
(2017)
\end{botherref}
\endbibitem

\bibitem[\protect\citeauthoryear{Guo et~al.}{2025}]{guo2025deepseek}
\begin{botherref}
\oauthor{\bsnm{Guo}, \binits{D.}},
\oauthor{\bsnm{Yang}, \binits{D.}},
\oauthor{\bsnm{Zhang}, \binits{H.}},
\oauthor{\bsnm{Song}, \binits{J.}},
\oauthor{\bsnm{Zhang}, \binits{R.}},
\oauthor{\bsnm{Xu}, \binits{R.}},
\oauthor{\bsnm{Zhu}, \binits{Q.}},
\oauthor{\bsnm{Ma}, \binits{S.}},
\oauthor{\bsnm{Wang}, \binits{P.}},
\oauthor{\bsnm{Bi}, \binits{X.}}, et al.:
Deepseek-r1: Incentivizing reasoning capability in llms via reinforcement learning.
arXiv preprint arXiv:2501.12948
(2025)
\end{botherref}
\endbibitem

\bibitem[\protect\citeauthoryear{Zellers et~al.}{2019}]{zellers2019defending}
\begin{botherref}
\oauthor{\bsnm{Zellers}, \binits{R.}},
\oauthor{\bsnm{Holtzman}, \binits{A.}},
\oauthor{\bsnm{Rashkin}, \binits{H.}},
\oauthor{\bsnm{Bisk}, \binits{Y.}},
\oauthor{\bsnm{Farhadi}, \binits{A.}},
\oauthor{\bsnm{Roesner}, \binits{F.}},
\oauthor{\bsnm{Choi}, \binits{Y.}}:
Defending against neural fake news.
Advances in neural information processing systems
\textbf{32}
(2019)
\end{botherref}
\endbibitem

\bibitem[\protect\citeauthoryear{Gehrmann et~al.}{2019}]{gehrmann2019gltr}
\begin{botherref}
\oauthor{\bsnm{Gehrmann}, \binits{S.}},
\oauthor{\bsnm{Strobelt}, \binits{H.}},
\oauthor{\bsnm{Rush}, \binits{A.M.}}:
Gltr: Statistical detection and visualization of generated text.
arXiv preprint arXiv:1906.04043
(2019)
\end{botherref}
\endbibitem

\bibitem[\protect\citeauthoryear{Wang et~al.}{2023}]{wang2023m4}
\begin{botherref}
\oauthor{\bsnm{Wang}, \binits{Y.}},
\oauthor{\bsnm{Mansurov}, \binits{J.}},
\oauthor{\bsnm{Ivanov}, \binits{P.}},
\oauthor{\bsnm{Su}, \binits{J.}},
\oauthor{\bsnm{Shelmanov}, \binits{A.}},
\oauthor{\bsnm{Tsvigun}, \binits{A.}},
\oauthor{\bsnm{Whitehouse}, \binits{C.}},
\oauthor{\bsnm{Afzal}, \binits{O.M.}},
\oauthor{\bsnm{Mahmoud}, \binits{T.}},
\oauthor{\bsnm{Sasaki}, \binits{T.}}, et al.:
M4: Multi-generator, multi-domain, and multi-lingual black-box machine-generated text detection.
arXiv preprint arXiv:2305.14902
(2023)
\end{botherref}
\endbibitem

\bibitem[\protect\citeauthoryear{Uchendu et~al.}{2021}]{uchendu2021turingbench}
\begin{botherref}
\oauthor{\bsnm{Uchendu}, \binits{A.}},
\oauthor{\bsnm{Ma}, \binits{Z.}},
\oauthor{\bsnm{Le}, \binits{T.}},
\oauthor{\bsnm{Zhang}, \binits{R.}},
\oauthor{\bsnm{Lee}, \binits{D.}}:
Turingbench: A benchmark environment for turing test in the age of neural text generation.
arXiv preprint arXiv:2109.13296
(2021)
\end{botherref}
\endbibitem

\bibitem[\protect\citeauthoryear{Mitchell et~al.}{2023}]{mitchell2023detectgpt}
\begin{bchapter}
\bauthor{\bsnm{Mitchell}, \binits{E.}},
\bauthor{\bsnm{Lee}, \binits{Y.}},
\bauthor{\bsnm{Khazatsky}, \binits{A.}},
\bauthor{\bsnm{Manning}, \binits{C.D.}},
\bauthor{\bsnm{Finn}, \binits{C.}}:
\bctitle{Detectgpt: Zero-shot machine-generated text detection using probability curvature}.
In: \bbtitle{International Conference on Machine Learning},
pp. \bfpage{24950}--\blpage{24962}
(\byear{2023}).
\bcomment{PMLR}
\end{bchapter}
\endbibitem

\bibitem[\protect\citeauthoryear{Hans et~al.}{2024}]{hans2024spotting}
\begin{botherref}
\oauthor{\bsnm{Hans}, \binits{A.}},
\oauthor{\bsnm{Schwarzschild}, \binits{A.}},
\oauthor{\bsnm{Cherepanova}, \binits{V.}},
\oauthor{\bsnm{Kazemi}, \binits{H.}},
\oauthor{\bsnm{Saha}, \binits{A.}},
\oauthor{\bsnm{Goldblum}, \binits{M.}},
\oauthor{\bsnm{Geiping}, \binits{J.}},
\oauthor{\bsnm{Goldstein}, \binits{T.}}:
Spotting llms with binoculars: Zero-shot detection of machine-generated text.
arXiv preprint arXiv:2401.12070
(2024)
\end{botherref}
\endbibitem

\bibitem[\protect\citeauthoryear{Kirchenbauer et~al.}{2023}]{kirchenbauer2023watermark}
\begin{bchapter}
\bauthor{\bsnm{Kirchenbauer}, \binits{J.}},
\bauthor{\bsnm{Geiping}, \binits{J.}},
\bauthor{\bsnm{Wen}, \binits{Y.}},
\bauthor{\bsnm{Katz}, \binits{J.}},
\bauthor{\bsnm{Miers}, \binits{I.}},
\bauthor{\bsnm{Goldstein}, \binits{T.}}:
\bctitle{A watermark for large language models}.
In: \bbtitle{International Conference on Machine Learning},
pp. \bfpage{17061}--\blpage{17084}
(\byear{2023}).
\bcomment{PMLR}
\end{bchapter}
\endbibitem

\bibitem[\protect\citeauthoryear{Zhao et~al.}{2023}]{zhao2023protecting}
\begin{bchapter}
\bauthor{\bsnm{Zhao}, \binits{X.}},
\bauthor{\bsnm{Wang}, \binits{Y.-X.}},
\bauthor{\bsnm{Li}, \binits{L.}}:
\bctitle{Protecting language generation models via invisible watermarking}.
In: \bbtitle{International Conference on Machine Learning},
pp. \bfpage{42187}--\blpage{42199}
(\byear{2023}).
\bcomment{PMLR}
\end{bchapter}
\endbibitem

\bibitem[\protect\citeauthoryear{Liu}{2019}]{liu2019roberta}
\begin{botherref}
\oauthor{\bsnm{Liu}, \binits{Y.}}:
Roberta: A robustly optimized bert pretraining approach.
arXiv preprint arXiv:1907.11692
\textbf{364}
(2019)
\end{botherref}
\endbibitem

\bibitem[\protect\citeauthoryear{Conneau et~al.}{2019}]{conneau2019unsupervised}
\begin{botherref}
\oauthor{\bsnm{Conneau}, \binits{A.}},
\oauthor{\bsnm{Khandelwal}, \binits{K.}},
\oauthor{\bsnm{Goyal}, \binits{N.}},
\oauthor{\bsnm{Chaudhary}, \binits{V.}},
\oauthor{\bsnm{Wenzek}, \binits{G.}},
\oauthor{\bsnm{Guzm{\'a}n}, \binits{F.}},
\oauthor{\bsnm{Grave}, \binits{E.}},
\oauthor{\bsnm{Ott}, \binits{M.}},
\oauthor{\bsnm{Zettlemoyer}, \binits{L.}},
\oauthor{\bsnm{Stoyanov}, \binits{V.}}:
Unsupervised cross-lingual representation learning at scale.
arXiv preprint arXiv:1911.02116
(2019)
\end{botherref}
\endbibitem

\bibitem[\protect\citeauthoryear{Li et~al.}{2014}]{li2014authorship}
\begin{bchapter}
\bauthor{\bsnm{Li}, \binits{J.S.}},
\bauthor{\bsnm{Monaco}, \binits{J.V.}},
\bauthor{\bsnm{Chen}, \binits{L.-C.}},
\bauthor{\bsnm{Tappert}, \binits{C.C.}}:
\bctitle{Authorship authentication using short messages from social networking sites}.
In: \bbtitle{2014 IEEE 11th International Conference on e-Business Engineering},
pp. \bfpage{314}--\blpage{319}
(\byear{2014}).
\bcomment{IEEE}
\end{bchapter}
\endbibitem

\bibitem[\protect\citeauthoryear{Horne et~al.}{2019}]{horne2019robust}
\begin{barticle}
\bauthor{\bsnm{Horne}, \binits{B.D.}},
\bauthor{\bsnm{N{\o}rregaard}, \binits{J.}},
\bauthor{\bsnm{Adali}, \binits{S.}}:
\batitle{Robust fake news detection over time and attack}.
\bjtitle{ACM Transactions on Intelligent Systems and Technology (TIST)}
\bvolume{11}(\bissue{1}),
\bfpage{1}--\blpage{23}
(\byear{2019})
\end{barticle}
\endbibitem

\bibitem[\protect\citeauthoryear{Guo et~al.}{2023}]{guo2023close}
\begin{botherref}
\oauthor{\bsnm{Guo}, \binits{B.}},
\oauthor{\bsnm{Zhang}, \binits{X.}},
\oauthor{\bsnm{Wang}, \binits{Z.}},
\oauthor{\bsnm{Jiang}, \binits{M.}},
\oauthor{\bsnm{Nie}, \binits{J.}},
\oauthor{\bsnm{Ding}, \binits{Y.}},
\oauthor{\bsnm{Yue}, \binits{J.}},
\oauthor{\bsnm{Wu}, \binits{Y.}}:
How close is chatgpt to human experts? comparison corpus, evaluation, and detection.
arXiv preprint arXiv:2301.07597
(2023)
\end{botherref}
\endbibitem

\bibitem[\protect\citeauthoryear{Xiong et~al.}{2024}]{xiong2024fine}
\begin{botherref}
\oauthor{\bsnm{Xiong}, \binits{F.}},
\oauthor{\bsnm{Markchom}, \binits{T.}},
\oauthor{\bsnm{Zheng}, \binits{Z.}},
\oauthor{\bsnm{Jung}, \binits{S.}},
\oauthor{\bsnm{Ojha}, \binits{V.}},
\oauthor{\bsnm{Liang}, \binits{H.}}:
Fine-tuning large language models for multigenerator, multidomain, and multilingual machine-generated text detection.
arXiv preprint arXiv:2401.12326
(2024)
\end{botherref}
\endbibitem

\bibitem[\protect\citeauthoryear{Wang et~al.}{2024}]{wang2024m4gt}
\begin{botherref}
\oauthor{\bsnm{Wang}, \binits{Y.}},
\oauthor{\bsnm{Mansurov}, \binits{J.}},
\oauthor{\bsnm{Ivanov}, \binits{P.}},
\oauthor{\bsnm{Su}, \binits{J.}},
\oauthor{\bsnm{Shelmanov}, \binits{A.}},
\oauthor{\bsnm{Tsvigun}, \binits{A.}},
\oauthor{\bsnm{Afzal}, \binits{O.M.}},
\oauthor{\bsnm{Mahmoud}, \binits{T.}},
\oauthor{\bsnm{Puccetti}, \binits{G.}},
\oauthor{\bsnm{Arnold}, \binits{T.}}, et al.:
M4gt-bench: Evaluation benchmark for black-box machine-generated text detection.
arXiv preprint arXiv:2402.11175
(2024)
\end{botherref}
\endbibitem

\bibitem[\protect\citeauthoryear{Li et~al.}{2023}]{li2023mage}
\begin{botherref}
\oauthor{\bsnm{Li}, \binits{Y.}},
\oauthor{\bsnm{Li}, \binits{Q.}},
\oauthor{\bsnm{Cui}, \binits{L.}},
\oauthor{\bsnm{Bi}, \binits{W.}},
\oauthor{\bsnm{Wang}, \binits{Z.}},
\oauthor{\bsnm{Wang}, \binits{L.}},
\oauthor{\bsnm{Yang}, \binits{L.}},
\oauthor{\bsnm{Shi}, \binits{S.}},
\oauthor{\bsnm{Zhang}, \binits{Y.}}:
Mage: Machine-generated text detection in the wild.
arXiv preprint arXiv:2305.13242
(2023)
\end{botherref}
\endbibitem

\bibitem[\protect\citeauthoryear{Dugan et~al.}{2024}]{dugan2024raid}
\begin{botherref}
\oauthor{\bsnm{Dugan}, \binits{L.}},
\oauthor{\bsnm{Hwang}, \binits{A.}},
\oauthor{\bsnm{Trhlik}, \binits{F.}},
\oauthor{\bsnm{Ludan}, \binits{J.M.}},
\oauthor{\bsnm{Zhu}, \binits{A.}},
\oauthor{\bsnm{Xu}, \binits{H.}},
\oauthor{\bsnm{Ippolito}, \binits{D.}},
\oauthor{\bsnm{Callison-Burch}, \binits{C.}}:
Raid: A shared benchmark for robust evaluation of machine-generated text detectors.
arXiv preprint arXiv:2405.07940
(2024)
\end{botherref}
\endbibitem

\bibitem[\protect\citeauthoryear{Lekkala et~al.}{2025}]{lekkala-etal-2025-cnlp}
\begin{bchapter}
\bauthor{\bsnm{Lekkala}, \binits{S.T.}},
\bauthor{\bsnm{Yadagiri}, \binits{A.}},
\bauthor{\bsnm{Vardhan}, \binits{M.S.}},
\bauthor{\bsnm{Pakray}, \binits{P.}}:
\bctitle{{CNLP}-{NITS}-{PP} at {G}en{AI} detection task 3: Cross-domain machine-generated text detection using {D}istil{BERT} techniques}.
In: \beditor{\bsnm{Alam}, \binits{F.}},
\beditor{\bsnm{Nakov}, \binits{P.}},
\beditor{\bsnm{Habash}, \binits{N.}},
\beditor{\bsnm{Gurevych}, \binits{I.}},
\beditor{\bsnm{Chowdhury}, \binits{S.}},
\beditor{\bsnm{Shelmanov}, \binits{A.}},
\beditor{\bsnm{Wang}, \binits{Y.}},
\beditor{\bsnm{Artemova}, \binits{E.}},
\beditor{\bsnm{Kutlu}, \binits{M.}},
\beditor{\bsnm{Mikros}, \binits{G.}} (eds.)
\bbtitle{Proceedings of the 1stWorkshop on GenAI Content Detection (GenAIDetect)},
pp. \bfpage{334}--\blpage{339}.
\bpublisher{International Conference on Computational Linguistics},
\blocation{Abu Dhabi, UAE}
(\byear{2025}).
\burl{https://aclanthology.org/2025.genaidetect-1.38/}
\end{bchapter}
\endbibitem

\bibitem[\protect\citeauthoryear{Wang et~al.}{2023}]{wang-etal-2023-seqxgpt}
\begin{bchapter}
\bauthor{\bsnm{Wang}, \binits{P.}},
\bauthor{\bsnm{Li}, \binits{L.}},
\bauthor{\bsnm{Ren}, \binits{K.}},
\bauthor{\bsnm{Jiang}, \binits{B.}},
\bauthor{\bsnm{Zhang}, \binits{D.}},
\bauthor{\bsnm{Qiu}, \binits{X.}}:
\bctitle{{S}eq{XGPT}: Sentence-level {AI}-generated text detection}.
In: \beditor{\bsnm{Bouamor}, \binits{H.}},
\beditor{\bsnm{Pino}, \binits{J.}},
\beditor{\bsnm{Bali}, \binits{K.}} (eds.)
\bbtitle{Proceedings of the 2023 Conference on Empirical Methods in Natural Language Processing},
pp. \bfpage{1144}--\blpage{1156}.
\bpublisher{Association for Computational Linguistics},
\blocation{Singapore}
(\byear{2023}).
\burl{https://aclanthology.org/2023.emnlp-main.73/}
\end{bchapter}
\endbibitem

\bibitem[\protect\citeauthoryear{Dugan et~al.}{2020}]{dugan-etal-2020-roft}
\begin{bchapter}
\bauthor{\bsnm{Dugan}, \binits{L.}},
\bauthor{\bsnm{Ippolito}, \binits{D.}},
\bauthor{\bsnm{Kirubarajan}, \binits{A.}},
\bauthor{\bsnm{Callison-Burch}, \binits{C.}}:
\bctitle{{R}o{FT}: A tool for evaluating human detection of machine-generated text}.
In: \beditor{\bsnm{Liu}, \binits{Q.}},
\beditor{\bsnm{Schlangen}, \binits{D.}} (eds.)
\bbtitle{Proceedings of the 2020 Conference on Empirical Methods in Natural Language Processing: System Demonstrations},
pp. \bfpage{189}--\blpage{196}.
\bpublisher{Association for Computational Linguistics},
\blocation{Online}
(\byear{2020}).
\doiurl{10.18653/v1/2020.emnlp-demos.25} .
\burl{https://aclanthology.org/2020.emnlp-demos.25/}
\end{bchapter}
\endbibitem

\bibitem[\protect\citeauthoryear{Kushnareva et~al.}{2023}]{kushnareva2023ai}
\begin{botherref}
\oauthor{\bsnm{Kushnareva}, \binits{L.}},
\oauthor{\bsnm{Gaintseva}, \binits{T.}},
\oauthor{\bsnm{Magai}, \binits{G.}},
\oauthor{\bsnm{Barannikov}, \binits{S.}},
\oauthor{\bsnm{Abulkhanov}, \binits{D.}},
\oauthor{\bsnm{Kuznetsov}, \binits{K.}},
\oauthor{\bsnm{Tulchinskii}, \binits{E.}},
\oauthor{\bsnm{Piontkovskaya}, \binits{I.}},
\oauthor{\bsnm{Nikolenko}, \binits{S.}}:
Ai-generated text boundary detection with roft.
arXiv preprint arXiv:2311.08349
(2023)
\end{botherref}
\endbibitem

\bibitem[\protect\citeauthoryear{Zeng et~al.}{2024}]{zeng2024towards}
\begin{botherref}
\oauthor{\bsnm{Zeng}, \binits{Z.}},
\oauthor{\bsnm{Liu}, \binits{S.}},
\oauthor{\bsnm{Sha}, \binits{L.}},
\oauthor{\bsnm{Li}, \binits{Z.}},
\oauthor{\bsnm{Yang}, \binits{K.}},
\oauthor{\bsnm{Liu}, \binits{S.}},
\oauthor{\bsnm{Ga{\v{s}}evi{\'c}}, \binits{D.}},
\oauthor{\bsnm{Chen}, \binits{G.}}:
Towards detecting ai-generated text within human-ai collaborative hybrid texts.
arXiv e-prints,
2403
(2024)
\end{botherref}
\endbibitem

\bibitem[\protect\citeauthoryear{Narayan et~al.}{2018}]{xsum-emnlp}
\begin{bchapter}
\bauthor{\bsnm{Narayan}, \binits{S.}},
\bauthor{\bsnm{Cohen}, \binits{S.B.}},
\bauthor{\bsnm{Lapata}, \binits{M.}}:
\bctitle{Don't give me the details, just the summary! {T}opic-aware convolutional neural networks for extreme summarization}.
In: \bbtitle{Proceedings of the 2018 Conference on Empirical Methods in Natural Language Processing},
\bconflocation{Brussels, Belgium}
(\byear{2018})
\end{bchapter}
\endbibitem

\bibitem[\protect\citeauthoryear{Zheng et~al.}{2015}]{zheng2015conditional}
\begin{bchapter}
\bauthor{\bsnm{Zheng}, \binits{S.}},
\bauthor{\bsnm{Jayasumana}, \binits{S.}},
\bauthor{\bsnm{Romera-Paredes}, \binits{B.}},
\bauthor{\bsnm{Vineet}, \binits{V.}},
\bauthor{\bsnm{Su}, \binits{Z.}},
\bauthor{\bsnm{Du}, \binits{D.}},
\bauthor{\bsnm{Huang}, \binits{C.}},
\bauthor{\bsnm{Torr}, \binits{P.H.}}:
\bctitle{Conditional random fields as recurrent neural networks}.
In: \bbtitle{Proceedings of the IEEE International Conference on Computer Vision},
pp. \bfpage{1529}--\blpage{1537}
(\byear{2015})
\end{bchapter}
\endbibitem

\bibitem[\protect\citeauthoryear{LaValley}{2008}]{lavalley2008logistic}
\begin{barticle}
\bauthor{\bsnm{LaValley}, \binits{M.P.}}:
\batitle{Logistic regression}.
\bjtitle{Circulation}
\bvolume{117}(\bissue{18}),
\bfpage{2395}--\blpage{2399}
(\byear{2008})
\end{barticle}
\endbibitem

\bibitem[\protect\citeauthoryear{Pal}{2005}]{pal2005random}
\begin{barticle}
\bauthor{\bsnm{Pal}, \binits{M.}}:
\batitle{Random forest classifier for remote sensing classification}.
\bjtitle{International journal of remote sensing}
\bvolume{26}(\bissue{1}),
\bfpage{217}--\blpage{222}
(\byear{2005})
\end{barticle}
\endbibitem

\bibitem[\protect\citeauthoryear{Chen et~al.}{2015}]{chen2015xgboost}
\begin{barticle}
\bauthor{\bsnm{Chen}, \binits{T.}},
\bauthor{\bsnm{He}, \binits{T.}},
\bauthor{\bsnm{Benesty}, \binits{M.}},
\bauthor{\bsnm{Khotilovich}, \binits{V.}},
\bauthor{\bsnm{Tang}, \binits{Y.}},
\bauthor{\bsnm{Cho}, \binits{H.}},
\bauthor{\bsnm{Chen}, \binits{K.}},
\bauthor{\bsnm{Mitchell}, \binits{R.}},
\bauthor{\bsnm{Cano}, \binits{I.}},
\bauthor{\bsnm{Zhou}, \binits{T.}}, \betal:
\batitle{Xgboost: extreme gradient boosting}.
\bjtitle{R package version 0.4-2}
\bvolume{1}(\bissue{4}),
\bfpage{1}--\blpage{4}
(\byear{2015})
\end{barticle}
\endbibitem

\bibitem[\protect\citeauthoryear{Balakrishnama and Ganapathiraju}{1998}]{balakrishnama1998linear}
\begin{barticle}
\bauthor{\bsnm{Balakrishnama}, \binits{S.}},
\bauthor{\bsnm{Ganapathiraju}, \binits{A.}}:
\batitle{Linear discriminant analysis-a brief tutorial}.
\bjtitle{Institute for Signal and information Processing}
\bvolume{18}(\bissue{1998}),
\bfpage{1}--\blpage{8}
(\byear{1998})
\end{barticle}
\endbibitem

\bibitem[\protect\citeauthoryear{Cervantes et~al.}{2020}]{cervantes2020comprehensive}
\begin{barticle}
\bauthor{\bsnm{Cervantes}, \binits{J.}},
\bauthor{\bsnm{Garcia-Lamont}, \binits{F.}},
\bauthor{\bsnm{Rodr{\'\i}guez-Mazahua}, \binits{L.}},
\bauthor{\bsnm{Lopez}, \binits{A.}}:
\batitle{A comprehensive survey on support vector machine classification: Applications, challenges and trends}.
\bjtitle{Neurocomputing}
\bvolume{408},
\bfpage{189}--\blpage{215}
(\byear{2020})
\end{barticle}
\endbibitem

\bibitem[\protect\citeauthoryear{Dessi et~al.}{2021}]{dessi2021tf}
\begin{botherref}
\oauthor{\bsnm{Dessi}, \binits{D.}},
\oauthor{\bsnm{Helaoui}, \binits{R.}},
\oauthor{\bsnm{Kumar}, \binits{V.}},
\oauthor{\bsnm{Recupero}, \binits{D.R.}},
\oauthor{\bsnm{Riboni}, \binits{D.}}:
Tf-idf vs word embeddings for morbidity identification in clinical notes: An initial study.
arXiv preprint arXiv:2105.09632
(2021)
\end{botherref}
\endbibitem

\bibitem[\protect\citeauthoryear{Allen and Hospedales}{2019}]{allen2019analogies}
\begin{bchapter}
\bauthor{\bsnm{Allen}, \binits{C.}},
\bauthor{\bsnm{Hospedales}, \binits{T.}}:
\bctitle{Analogies explained: Towards understanding word embeddings}.
In: \bbtitle{International Conference on Machine Learning},
pp. \bfpage{223}--\blpage{231}
(\byear{2019}).
\bcomment{PMLR}
\end{bchapter}
\endbibitem

\bibitem[\protect\citeauthoryear{Luan and Lin}{2019}]{luan2019research}
\begin{bchapter}
\bauthor{\bsnm{Luan}, \binits{Y.}},
\bauthor{\bsnm{Lin}, \binits{S.}}:
\bctitle{Research on text classification based on cnn and lstm}.
In: \bbtitle{2019 IEEE International Conference on Artificial Intelligence and Computer Applications (ICAICA)},
pp. \bfpage{352}--\blpage{355}
(\byear{2019}).
\bcomment{IEEE}
\end{bchapter}
\endbibitem

\bibitem[\protect\citeauthoryear{Liu et~al.}{2016}]{liu2016recurrent}
\begin{botherref}
\oauthor{\bsnm{Liu}, \binits{P.}},
\oauthor{\bsnm{Qiu}, \binits{X.}},
\oauthor{\bsnm{Huang}, \binits{X.}}:
Recurrent neural network for text classification with multi-task learning.
arXiv preprint arXiv:1605.05101
(2016)
\end{botherref}
\endbibitem

\bibitem[\protect\citeauthoryear{Zhou et~al.}{2015}]{zhou2015c}
\begin{botherref}
\oauthor{\bsnm{Zhou}, \binits{C.}},
\oauthor{\bsnm{Sun}, \binits{C.}},
\oauthor{\bsnm{Liu}, \binits{Z.}},
\oauthor{\bsnm{Lau}, \binits{F.}}:
A c-lstm neural network for text classification.
arXiv preprint arXiv:1511.08630
(2015)
\end{botherref}
\endbibitem

\bibitem[\protect\citeauthoryear{Tie et~al.}{2025}]{tie2025research}
\begin{barticle}
\bauthor{\bsnm{Tie}, \binits{R.}},
\bauthor{\bsnm{Li}, \binits{M.}},
\bauthor{\bsnm{Zhou}, \binits{C.}},
\bauthor{\bsnm{Ding}, \binits{N.}}:
\batitle{Research on the application of an improved autoformer model integrating cnn-attention-bigru in short-term power load forecasting}.
\bjtitle{Evolving Systems}
\bvolume{16}(\bissue{3}),
\bfpage{98}
(\byear{2025})
\end{barticle}
\endbibitem

\bibitem[\protect\citeauthoryear{Hu and Zhao}{2021}]{hu2021bi}
\begin{barticle}
\bauthor{\bsnm{Hu}, \binits{Y.-L.}},
\bauthor{\bsnm{Zhao}, \binits{Q.-S.}}:
\batitle{Bi-gru model based on pooling and attention for text classification}.
\bjtitle{International Journal of Wireless and Mobile Computing}
\bvolume{21}(\bissue{1}),
\bfpage{26}--\blpage{31}
(\byear{2021})
\end{barticle}
\endbibitem

\bibitem[\protect\citeauthoryear{Devlin}{2018}]{devlin2018bert}
\begin{botherref}
\oauthor{\bsnm{Devlin}, \binits{J.}}:
Bert: Pre-training of deep bidirectional transformers for language understanding.
arXiv preprint arXiv:1810.04805
(2018)
\end{botherref}
\endbibitem

\bibitem[\protect\citeauthoryear{Sanh}{2019}]{sanh2019distilbert}
\begin{botherref}
\oauthor{\bsnm{Sanh}, \binits{V.}}:
Distilbert, a distilled version of bert: smaller, faster, cheaper and lighter.
arXiv preprint arXiv:1910.01108
(2019)
\end{botherref}
\endbibitem

\bibitem[\protect\citeauthoryear{He et~al.}{2020}]{he2020deberta}
\begin{botherref}
\oauthor{\bsnm{He}, \binits{P.}},
\oauthor{\bsnm{Liu}, \binits{X.}},
\oauthor{\bsnm{Gao}, \binits{J.}},
\oauthor{\bsnm{Chen}, \binits{W.}}:
Deberta: Decoding-enhanced bert with disentangled attention.
arXiv preprint arXiv:2006.03654
(2020)
\end{botherref}
\endbibitem

\bibitem[\protect\citeauthoryear{Warner et~al.}{2024}]{warner2024smarter}
\begin{botherref}
\oauthor{\bsnm{Warner}, \binits{B.}},
\oauthor{\bsnm{Chaffin}, \binits{A.}},
\oauthor{\bsnm{Clavi{\'e}}, \binits{B.}},
\oauthor{\bsnm{Weller}, \binits{O.}},
\oauthor{\bsnm{Hallstr{\"o}m}, \binits{O.}},
\oauthor{\bsnm{Taghadouini}, \binits{S.}},
\oauthor{\bsnm{Gallagher}, \binits{A.}},
\oauthor{\bsnm{Biswas}, \binits{R.}},
\oauthor{\bsnm{Ladhak}, \binits{F.}},
\oauthor{\bsnm{Aarsen}, \binits{T.}}, et al.:
Smarter, better, faster, longer: A modern bidirectional encoder for fast, memory efficient, and long context finetuning and inference.
arXiv preprint arXiv:2412.13663
(2024)
\end{botherref}
\endbibitem

\end{thebibliography}
\end{document}